\documentclass[conference]{IEEEtran}
\makeatletter
\newcounter{parentalgorithm}
\newenvironment{subalgorithms}{%
  \refstepcounter{algorithm}%
  \protected@edef\theparentalgorithm{\thealgorithm}%
  \setcounter{parentalgorithm}{\value{algorithm}}%
  \setcounter{algorithm}{0}%
  \def\thealgorithm{\theparentalgorithm.\arabic{algorithm}}%
  \ignorespaces
}{%
  \setcounter{algorithm}{\value{parentalgorithm}}%
  \ignorespacesafterend
}
\makeatother

\IEEEoverridecommandlockouts %To add sponsors and index terms

\usepackage{cite}
\usepackage{textpos}
\usepackage{algorithm,algorithmic}
\usepackage{makecell}
\ifCLASSINFOpdf
   \usepackage[pdftex]{graphicx}
   \graphicspath{{./Images/}}
   \DeclareGraphicsExtensions{.pdf,.jpeg,.png}
\else
\fi
\usepackage[cmex10]{amsmath}
\usepackage{multirow}
\usepackage{booktabs,colortbl}

  \renewcommand\footnoterule{\vspace*{-3pt}%
     \hrule width 2in height 0.4pt
     \vspace*{2.6pt}}

\makeatletter
\newcommand{\bs}[1]{\boldsymbol{#1}}

\newcommand{\Rmnum}[1]{\expandafter\@slowromancap\romannumeral #1@}
\makeatother
\DeclareMathOperator*{\argmax}{arg\,max}

\begin{document}

\title{Reinforcement Learning for \\
the Unit Commitment Problem}

\author{%

\IEEEauthorblockN{Gal Dalal \\ Shie Mannor }
\IEEEauthorblockA{Department of Electrical Engineering\\
Technion\\
Haifa, Israel\\
gald@tx.technion.ac.il\\
shie@ee.technion.ac.il}

}


\maketitle

%TODO(if have space): add response paragraph to review comments; split intro into subsections; split MDP model into subsections (or items);inflate experiments section;inflate summary
%ABSTRACT
\begin{abstract}
In this work we solve the day-ahead unit commitment (UC) problem, by formulating it as a Markov decision process (MDP) and finding a low-cost policy for generation scheduling. We present two reinforcement learning algorithms, and devise a third one. We compare our results to previous work that uses simulated annealing (SA), and show a 27\% improvement in operation costs, with running time of 2.5 minutes (compared to 2.5 hours of existing state-of-the-art). \\
\end{abstract}

%INDEX TERMS
\begin{IEEEkeywords}
Power generation dispatch, Learning (artificial intelligence), Optimal scheduling, Optimization methods.
\end{IEEEkeywords}

\section{Introduction}
Unit commitment (UC) is the process of determining the most cost-effective combination of generating units and their generation levels within a power
system to meet forecasted load and reserve requirements, while adhering to generator and transmission constraints \cite{UC_survey}. 
This is a non-linear, mixed-integer combinatorial optimization problem \cite{UC_synopsis}. Low-cost solutions to this problem will directly translate into low production costs for power utilities.
As the size of the problem increases, it becomes a very complex, hard to solve problem \cite{UC_ls}. \\
 Multiple optimization approaches have been applied over the past years, %change a bit
 such as the priority ordering methods \cite{PO1,PO2}, dynamic programming \cite{DP}, Lagrangian relaxation \cite{LR1}, the branch-and-bound method \cite{BAB}, and the integer and mixed-integer programming \cite{MILP1,MILP2}.
 Other, more recent methods are from the field of artificial intelligence, such as the expert systems \cite{ES1}, neural networks \cite{NN}, fuzzy logic \cite{FL}, genetic algorithms \cite{GA}, and simulated annealing \cite{SA}. \\
 Many of these approaches are either purely heuristic (e.g. priority ordering) or semi-heuristic (e.g. simulated annealing) , thus are often very sensitive to choice of architecture, manual parameter tuning, and different cost functions. On the other hand, analytical methods can also introduce critical shortcomings. The branch-and-bound algorithm, for instance, suffers from an exponential growth in execution time with the size of the UC problem \cite{BAB,LR3}. In addition, using approximations for making it tractable for large scale systems causes solutions to be highly sub-optimal. \\
 Therefore in our work,  we take an analytical approach to the problem, while assuring it will not become intractable nor highly suboptimal in large scale systems. We use a Markov Decision Process (MDP) framework. MDPs are used to describe numerous phenomena in many fields of science \cite{MDP}. Such a model is aimed to describe a  decision making process, where  outcomes of the process are partly random and partly under the control of the decision maker. \\
In this work we assume that generation cost functions of the different generators are known to the decision maker. We note that this is often not the case with European system operators, since in a European competitive electricity market, cost information is not available. However, in many other cases this information is indeed available, such as in some north American TSOs, and generation companies with multiple generation units (such a company would not know the characteristics of the power system, nevertheless it is not problematic since they do not play a role in our formulation). In addition, the UC problem can easily
be extended to generate production schedules in a competitive
market environment\cite{uc_market1}. Another paper shows the framework in which a
traditional cost-based unit commitment tool can be used to
assist bidding strategy decisions to a day-ahead electricity
pool market \cite{uc_market2}.
In general, European TSOs can approximate generation costs based on historical data (that include past and present bids they receive from generators) and market simulation. Also, in future work, the uncertainty in these approximations can be  naturally expressed in our MDP model. \\
The rest of the paper is organized as follows. Section \Rmnum{2} formulates the unit commitment problem. We then present our MDP model for the UC problem in section \Rmnum{3} , and give an introduction to reinforcement learning in section \Rmnum{4}. The algorithms we use are presented in section \Rmnum{5}. Then, in section \Rmnum{6} we show numerical tests of our methods. Lastly, in section \Rmnum{7} we summarize our work.

\section{Unit Commitment Problem Formulation}
The problem is formulated as the following constrained optimization program. 
\subsection{Objective}
The objective is to find a feasible plan with minimal cost for operating generators to meet client demands --
\begin{eqnarray} \label{obj}
\min_{\alpha_{i}(t),P_{i}(t),\forall i,t}\sum_{t=1}^{T}\sum_{i=1}^{N}[\alpha_{i}(t)C_{i}(P_{i}(t))&+&\\ \nonumber
\alpha_{i}(t)[1-\alpha_{i}(t-1)]SC_{i}(t_{off_{i}})].&&
\end{eqnarray}

%\begin{eqnarray} 
%&b&c\\
%a2&b1&c1
%\end{eqnarray}

Where:\\
$a_{i}(t)=1$ when unit $i$ is turned on at time $t_{i}$, and $\alpha_{i}(t)=0$ otherwise. \\
$P_{i}(t)$ is the injected power $[MW]$ in unit $i$ at time $t$.\\
$C_{i}(P)$ is the cost $[\$]$ of injecting power $P$ in unit $i$ .\\
$SC_i(t_{off_i})$ is the start-up cost $[\$]$ of unit $i$ after it has been off for a time period of $t_{off_i}$.

\subsection{Constraints}
Any feasible solution is subject to the following constraints:
\begin{itemize}
\item	Load balance --
\begin{equation}
\forall t:\sum_{i=1}^{N}(\alpha_{i}(t)P_{i}(t))=D(t).
\end{equation} 
\item Generation limits --
\begin{equation}
\forall i,t:\alpha_{i}(t)P_{min_{i}}\leq P_{i}(t)\leq\alpha_{i}(t)P_{max_{i}}.
\end{equation} 
\item Set generation limits --
\begin{eqnarray}
\forall t:(\sum_{i=1}^{N}\alpha_{i}(t)P_{min_{i}})&\leq &D(t),\\ \nonumber
(\sum_{i=1}^{N}\alpha_{i}(t)P_{max_{i}})&\geq &D(t)+R(t).
\end{eqnarray}
\item Minimum up/down time --
\begin{eqnarray}
\forall i:t_{off_{i}}&\geq &t_{down_{i}},\\ \nonumber
t_{on_{i}}&\geq &t_{up_{i}}.
\end{eqnarray}
\end{itemize}

Where: \\
$D(t)$ is the demand at time $t$. \\
$R(t)$ is the needed power reserve at time $t$. \\
$P_{min_{i}},P_{max_{i}}$ are the minimal and maximal power injections
in unit $i$. \\
$t_{off_{i}},t_{on_{i}}$ are the minimal time periods before turning unit $i$ on/off.

\subsection{Costs}

\begin{figure}[!ht] 
\centering
\includegraphics[scale=0.2]{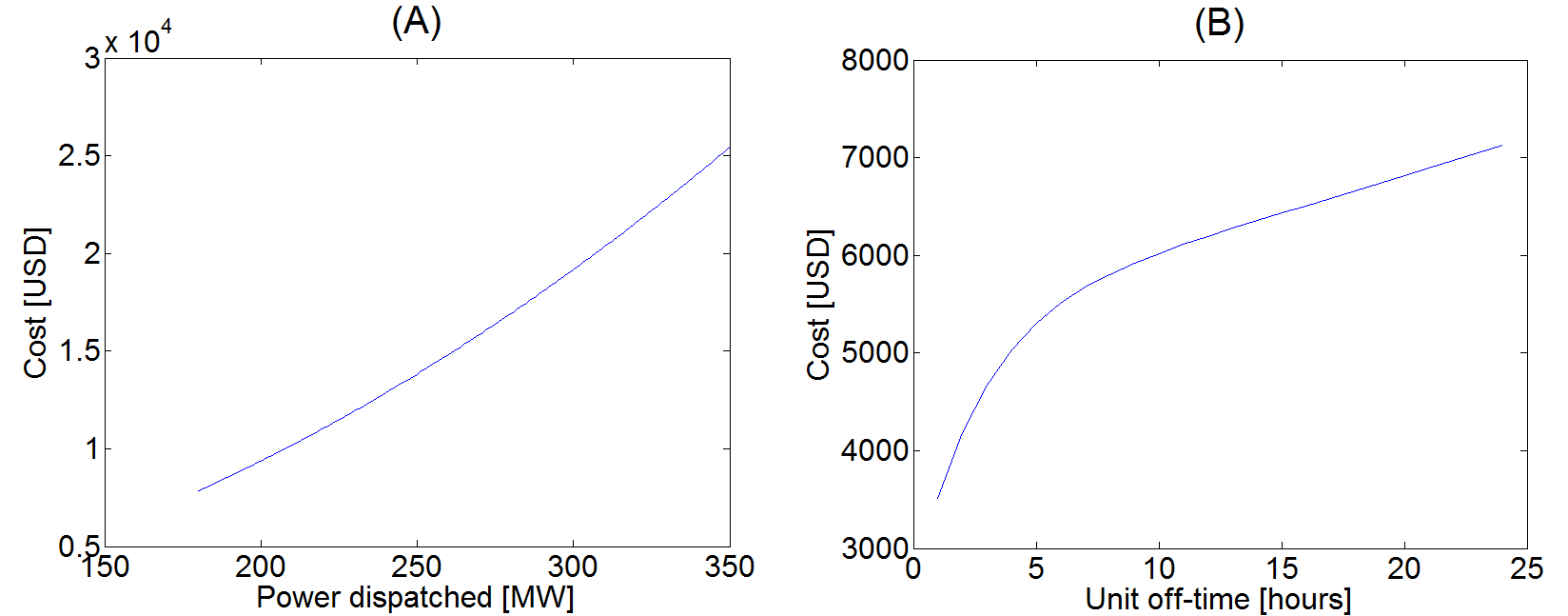}
\caption{(A) shows the generation cost of a specific generator, and (B) shows the start-up cost of that generator, as a function of the time it was off}
\label{fig:costs}
\end{figure}

\begin{itemize}
\item Generation Cost -- quadratic function of the power generated by that
unit:%\vspace{0.4cm}

\begin{equation}
C_{i}(P_{i})=a_{i}P_{i}^{2}+b_{i}P_{i}+c_{i}.
\end{equation}
%\vspace{1cm}
%\begin{textblock*}{\textwidth}(7.67cm,-19.5mm)    
%\includegraphics[scale=0.45]{fuel_cost}
%\end{textblock*}

\item Start-up cost -- an exponentially dependent function of
the number of hours a unit has been down: %\vspace{0.4cm}

\begin{equation}
SC_{i}(t_{off_{i}})=e_{i}\exp(-g_{i} t_{off_{i}})+f_{i}\exp(-h_{i} t_{off_{i}}).
\end{equation}
%\begin{textblock*}{\textwidth}(7.69cm,-10.5mm)    
%\includegraphics[scale=0.4]{SU_cost}
%\end{textblock*}

\end{itemize}
A graphical example of generation (A) and start-up (B) costs of a specific generator (with the parameters used in the experiments section) is displayed in Figure \ref{fig:costs}. 

\section{Markov Decision Process Approach}
Finding a global optimum is intractable for  this non convex, mixed integer-quadratic problem. Therefore, unlike in \cite{SA}, where a direct search in the solution space was performed, we suggest an alternative approach: decomposing the objective into a sequential decision making process.
We use a Markov Decision Process 4-tuple $(S,A,P,R)$ \cite{MDP} to model the system's dynamics. Briefly, in this model at each time step, the process is in some state $s$, an action $a$ is taken, the system transitions to a next state $s'$ according to a transition kernel $P(s,a,s')$, and a reward $R(s,a,s')$ is granted. Thus, the next state $s'$ depends only on the current state $s$ and the decision maker's action $a$.\\
\subsection{State-Space}
The system's state can be fully described by the on/off time of each of the $N$ generators, and the time of the day (negative values indicate off time):
\begin{equation} \nonumber
S=\{-24,-23,\dots,-1,1,2,\dots,24\}^{N}\times\{1,\dots,24\}.
\end{equation}

\subsection{Action Space}
Each unit can be turned/kept on, or turned/kept off:
\[A=\{0,1\}^{N}.\]
\subsection{Reward}
At each time step, the reward is (minus) the cost of operation of the $N$ machines:
\begin{equation} \nonumber
R(s,a,s')=-\sum_{i=1}^{N}[I_{[s'_{i}>0]}C_{i}(P_{i})+I_{[s'_{i}>0]}I_{[s_{i}<0]}SC_{i}(s_{i})].
\end{equation}
The power injections $P_{i}$ are chosen by solving the appropriate constrained quadratic program (generation cost is quadratic). \\
By maximizing the undiscounted cumulative reward of the MDP, we minimize
the original objective.
\subsection{Transition Kernel}
\begin{figure}[!ht] 
\centering
\includegraphics[scale=0.28]{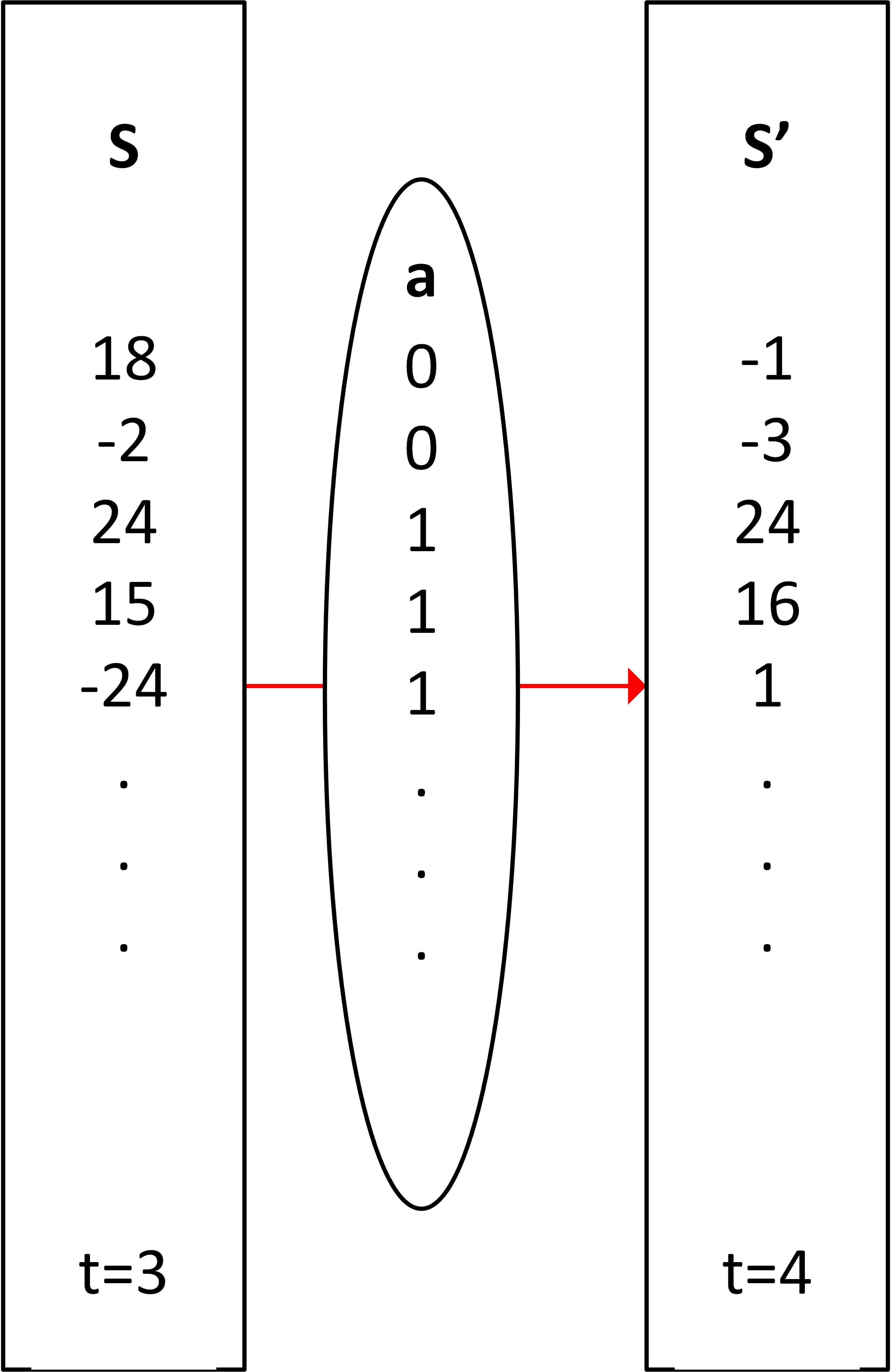}
\caption{Example for state transition - $f(s,a)=s'$. Generators are turned/kept on or off when an action of 1 or 0 is taken. Time is also represented in the state.}
\label{fig:trans}
\end{figure}

Transition is deterministic:
$f(s,a)=s'$. The transition function restricts the process to satisfy the constraints of the optimization problem.\\ A transition example is presented in Figure \ref{fig:trans}.

\section{Reinforcement Learning}
A \textit{policy} is a mapping between a state-space $S$ and an action-space $A$. Given a policy, we know what action to perform at each state of the system.\\
For the defined MDP, our goal is the following:
Find an optimal policy $\pi^{*}:S\rightarrow A$ s.t:
\begin{equation} \label{OP}
\pi^{*}=\arg\max{}_{\pi\in\Pi}\sum_{t=1}^{T}R(s_{t},\pi(s_{t}),f(s_{t},\pi(s_{t}))).
\end{equation}

Where $\Pi$ is the space of all possible policies.\\
\textit{Reinforcement Learning} (RL) \cite{RL} is a field in Machine Learning that studies algorithms that learn by interacting with the environment. 
The learning is done for states and actions. For each state $s$, given a policy $\pi$, a \textit{state value} $v^{\pi}(s)$ is defined as:
\begin{equation}
v^{\pi}(s)=\sum_{t=1}^{T}R(s_{t},\pi(s_{t}),f(s_{t},\pi(s_{t})))\quad \textrm{for} \quad s_{1}=s.
\end{equation}
\section{Reinforcement Learning Algorithmic Solutions}
In this section we present three different reinforcement learning algorithms for solving \ref{OP}.
\subsection{Algorithm 1 -- Approximate Policy Iteration (API) using Classification}
%API \cite {API} is an algorithm that iterates between two stages: regression-based estimation of the states' values under a fixed policy, and the improvement of the policy using the learned values.
%In the full paper we will explain our implementation of the API algorithm with policy as a classifier, and the reason it did not scale up to the full problem presented in the 'experiments' section, but only up to a smaller setting.
An extension to the state value $v^{\pi}(s)$ defined above, is the \textit{state-action value} function $Q^{\pi}(s,a)$, which denotes the value of performing action $a$ (regardless of the policy $\pi$), and only after that -- following policy $\pi$. In our first algorithm we use the state-action value function. This function is defined the set of all $(s,a)$ pairs, which is of size $|S|\cdot |A|$.\\
\subsubsection{Approximation}
Our state-space grows exponentially with $N$: $|S|=24\cdot 48^{N}$, as well as the action-space: $|A|=2^N$. Already for $N\geq 4$, it is impossible to find the exact value for each state-action pair. We therefore use an approximation method for evaluating the state-action value function $Q^{\pi}(s,a)$. We use feature-based regression, which significantly lowers the dimension of $Q^\pi(\cdot,\cdot)$ from $|S|\cdot |A|$ to $dim(\phi(s,a))$, the dimension of the feature vector $\phi(s,a)$. We use 4 binary features for each generator $i$, for each of the possible 'interesting' zones it can be in:
\begin{eqnarray}
&s_i&<-t_{off_i}, \nonumber \\ \nonumber -t_{off_i}\leq &s_i&<0, \\ \nonumber 0 < &s_i& \leq t_{on_i}, \\ \nonumber t_{on_i}<&s_i& .\nonumber
\end{eqnarray}
The features are then duplicated $N$ times and zeroed out at indices where the action vector is $0$. The result is again duplicated into two -- for distinguishing between $(s,a)$ pairs that will lead to a catastrophe (transition to infeasible states). We end up with a feature vector $\phi(s,a)$ of dimension that is only quadratic in $N$: $dim(\phi(s,a))=2\cdot 4\cdot N^2$.
\subsubsection{Policy Iteration Algorithm}
The basic algorithm is policy iteration \cite{RL}. This well-known algorithm iterates between two stages: evaluation of the states' values under a fixed policy, and the improvement of the policy using the learned values. We perform the evaluation using the SARSA \cite{RL} algorithm (with epsilon-greedy exploration), and the improvement is simply done using the following maximization (for step $k$): 
\begin{equation}
\pi_{k}(s)=\argmax_{a\in A}Q_k(s,a)=\argmax_{a\in A}\phi(s,a)^{T}w_k.
\end{equation}
\subsubsection{Policy Representation}
The biggest challenge in enabling policy iteration for our problem is the choice of policy representation. On the one hand, the policy should be defined for all states $s\in S$. On the other hand, it can practically only be trained using a very small fraction of this huge state-space. Also, its output is a selection from the enormous space of actions. \\
To handle the above difficulties, we chose the policy to be a classifier, that classifies states into actions. This gives rise to a large-scale multi-class classification task ($2^N$ optional classes), which is considered to be a difficult problem on its own. We tackle that by using a hierarchical classifier with a tree-based structure: Each node classifies an action bit and splits into two nodes for the next bit. Classification is done in the feature-space. The perceptron algorithm \cite{perceptron} is used a the basic binary classifier (online updates can be made to save memory). A different tree is stored per each time-step. \\Note that this is not a decision-tree classifier, but multiple binary hyper-plane based classifiers that are being traversed through in a sequential manner. The leafs determine the final action prediction (encapsulate the path).	

\begin{algorithm}[H]
Initialize:\\
$\alpha$ - SARSA step size\\
$\epsilon$ - exploration parameter\\
$N_{pi}$ - iteration count\\
$\pi_0$ - intial policy\\
$\phi$ - basis functions
\begin{algorithmic}[1]
\FOR{$k=1$ to $k=N_{pi}$}
\STATE $w_{k}=SARSA(\pi_{k},\alpha,\epsilon)$
\FOR{all $s\in S$}
\STATE $a^*=\argmax_{a\in A}\phi(s,a)^{T}w_k$
\STATE $\pi_{k}=updateClassifier(a^*,\pi_{k})$
\ENDFOR
\ENDFOR
\RETURN $\pi_{N_{pi}}$
\end{algorithmic} 
\caption{Approximate Policy Iteration using Classification} 
\label{alg:seq} 
\end{algorithm} 
\subsection{Algorithm 2 -- Tree Search}
An MDP can be represented as a tree, where each node is a state and each edge corresponds to an action. In our problem, we can theoretically express the tree explicitly, where the edges include the exact reward of the transition since transitions and rewards are deterministic. Let us also denote $s^t_j$ so be the $j$-th state at time-step $t$.
\begin{figure}[!ht] 
\centering
\includegraphics[scale=0.45]{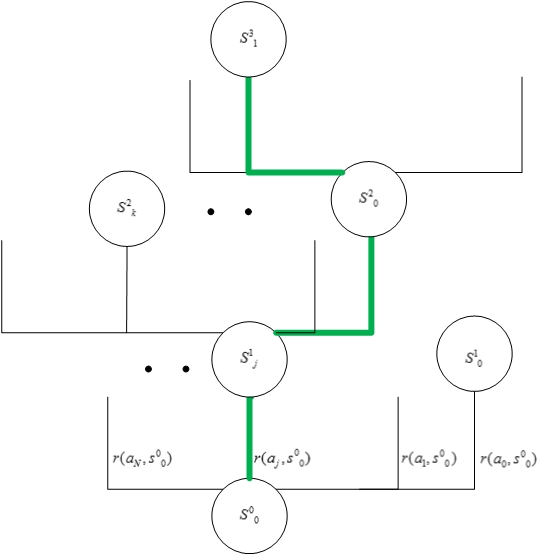}
\caption{Visualisation of algorithm 2 -- tree search. Nodes are states, edges are transitions with the corresponding rewards.}
\label{fig:tree}
\end{figure}
Under this representation, finding an optimal policy $\pi^*$ corresponds to finding the largest aggregated reward path from the root (initial state $s^0_0$).\\ 
However, since the number of possible paths in the tree is in the order of $|A|^T=2^{N\cdot T}$ , naively searching the tree for an optimal path is intractable for a problem of our size. Therefore in our tree search algorithm we limit the time horizon in which we search to be $H$ ($H<T$). I.e, tree search seeks a lookahead policy by iterating through all of the possible outcomes in a limited lookahead horizon $H$. Our algorithm contains the two following main components:
\subsubsection{Algorithm 2.1}
The first part of the algorithm, $findBestAction(s^t,H)$, recursively searches for the next best action that can be taken from state $s^t$, by iterating on all possible actions from that state. "Best" in this case, is considered with regard to all possible paths in a lookahead horizon of $H$ time-steps ahead. That is, it finds $a^t$, the first action in the vector $\bs{a}=(a^t,a^{t+1},\dots,a^{t+H})$, where \[\bs{a}=\argmax_{\bs{a'}\in A^H} \sum_{t'=t}^{t+H}R(s^{t'},a^{t'},f(s^{t'},a^{t'})).\]

\setcounter{algorithm}{1}
\begin{subalgorithms}
\begin{algorithm}[H]
\begin{algorithmic}[1] 
\IF{$ H==0 \mbox{ } \OR \mbox{ }  t==T-1 $}
\RETURN $(0,0)$
\ENDIF
\STATE $v_{m}=-\infty,a_{m}=\bar{0}$
\FOR{all $a\in A$}
\STATE $s^{t+1}=f(s^{t},a)$
\STATE $(a^{t+1},v^{t+1})=findBestAction(s^{t+1},H-1)$
\STATE $v^{t}=R(s^{t},a,s^{t+1})+v^{t+1}$
\IF{$v^{t}\geq v_{m}$}
\STATE $a_{m}=a,v_{m}=v$
\ENDIF
\ENDFOR
\RETURN $(a_{m},v_{m})$
\end{algorithmic} 
\caption{$(a_{m},v_{m})=findBestAction(s^{t},H)$ } 
\label{alg:seq} 
\end{algorithm} 
\subsubsection{Algorithm 2.2}
The second component of our tree search algorithm finds the optimal lookahead policy  by initiating $findBestPolicy$ per each time step from $t=0$ to $t=T-1$.

\begin{algorithm}[H]
\begin{algorithmic}[1] 
\FOR{$t=0$ to $t=T-1$}
\STATE $(a_{m},v_{m})=findBestAction(s^{t},H)$
\STATE $\pi(s^{t})=a_{m}$
\STATE $s^{t+1}=f(s^{t},a_{m})$
\ENDFOR
\RETURN $\pi$
\end{algorithmic} 
\caption{$\pi=treeSearch(s^{t},H)$ } 
\end{algorithm} 
\end{subalgorithms}
\subsubsection{Improve by Sub-sampling}
We can take advantage of a certain property of this problem: it is very unlikely that in "good" (highly rewarding) paths, subsequent actions will differ significantly from each other. This is both because of the high start-up costs of generators, and because of the minimal up/down time limitation (rapidly switching different machines on/off can lead to infeasible states, where there aren't enough available generators to satisfy demand).\\
Exploiting this property, we added an improvement for our algorithm. Instead of iterating throughout all actions when searching for the best one at time $t+1$, we only sample small deviations from last best action at time $t$ (denoted as $a^{t*}$). For the sampling we use a probability density over the action $a^{t+1}$, with an inverse relation to $\|a^{t*}-a^{t+1}\|_2$. 
%When searching for an take advantage of the small changes in the action vector
This improvement significantly reduces the runtime, and can also enable setting a larger value for $H$.
In the experiments section, we test the usability of our improvement and compare it to the original approach. 
\subsection{Algorithm 3 -- Back Sweep}
Our "Back Sweep" algorithm is a novel algorithm, inspired by the concept in dynamic programming of backtracking from the terminal time and going backwards. That way, we have a reliable estimation of the value of future states, and can base decisions correctly based on that knowledge of future values. The main novelty is in sampling 'interesting' (potentially beneficial) areas of the state-space, and use a nearest neighbour (NN) approximation of them in the Bellman update step (defined below).

\subsubsection{Algorithm 3.1}
First of two parts of the algorithm is evaluation the optimal value of each sampled state, $v^*(s)$. The optimal value is the value of states when using the optimal policy as defined in \ref{OP}:
\begin{equation}
v^*(s)=\sup \limits_{\pi \in \Pi} v^\pi(s).
\end{equation}
It is found using Bellman's update step, that lies in the heart of the algorithm:

\begin{equation}
v^{*}(s)=\max \limits_{a \in A} [R(s,a,s') + v^{*}(s')].
\end{equation}
\setcounter{algorithm}{2}
\begin{subalgorithms}

\begin{algorithm}[H]
\begin{algorithmic}[1] 
\STATE Initialize ${\cal D}=\emptyset,\tilde{s}=s^{T}$
\FOR{$t=T-1$ to $0$} 
\STATE $\underbar{S}^{t} =$  $ sampleEnvironment(\tilde{s},N_{s})$ 
\FOR{$i=1$ to $N_{s}$} 
\STATE $\ensuremath{\hat{v}}^{*t}(s^{t}_{i}) = \max_{a\in A}[R(s^{t}_{i},a,f(s^{t}_{i},a))+\hat{v}^{*t+1}(NN(f(s^{t}_{i},a),\cal D))]$ 
\ENDFOR 
\STATE ${\cal {\cal D}=D}\cup\{(\underbar{S}^{t},\hat{\underbar{V}}^{*t})\}$
\STATE $\tilde{s}=\arg\max_{s}\hat{\underbar{V}}^{*t}(s)$
\ENDFOR 
\RETURN $\cal D$
\end{algorithmic} 
\caption{${\cal D}=evaluateStates(N_{s},s_{0})$} 
\end{algorithm}

\begin{itemize}
\item $sampleEnvironment(\tilde{s},N_{s})$ returns $N_{s}$ samples of
states that are 'close' to $\tilde{s}$. Closeness is quantified using a metric we defined.
\item $NN(s,\cal D)$ returns the nearest-neighbor state from all states that
are in the $(s,v)$ pairs in $\cal D$.
\end{itemize}

\subsubsection{Algorithm 3.2}
The second part of the algorithm will produce a greedy policy via one quick sweep forward, beginning from the initial state $s^0$. This policy is greedy since at each step we choose the best possible action, and it is proven that for exact $v^*$ values, it will also be the optimal policy \cite{RL}.
\begin{algorithm}[H]
\begin{algorithmic}[1] 
\FOR{$t=0$ to $T-1$} 
\STATE $a^{t}=\arg\max_{a\in A}[R(s^{t},a,f(s^{t},a))+\hat{v}^{*t+1}(NN(f(s^{t}_{i},a),\cal D))]$ 
\STATE $\pi(s^{t})=a^{t}$
\STATE $s^{t+1}=f(s^{t},a^{t})$
\ENDFOR 
\RETURN $\pi$
\end{algorithmic} 
\caption{$\pi=findGreedyPolicy({\cal D})$ } 
\end{algorithm} 
\end{subalgorithms}

\section{Experiments}
In order to test the performance of the three proposed algorithms, we used Matlab \cite{matlab} to implement and run them on a problem setting with $N=12$ generators, a 24-hour schedule ($T=24$), with parameters taken from \cite{SA}. In \cite{SA}, an Adaptive Simulated Annealing (SA) technique is used, and a minimal objective of \$644,951 is achieved.
\subsection{Algorithm 1}
Algorithm 1 only performed well on a smaller setting of the problem ($N=8$, $T=12$) and was not included in Table \ref{tab:comparison}. 
In spite of that, we chose to present algorithm 1 in this paper as a baseline. API is a very commonly used algorithm in the reinforcement learning literature. On top of that, the policy structure we devised enabled the algorithm the leap from performing only on a $N=4$, $T=8$ setting, to the $N=8$, $T=12$ setting.\\
We also find value in understanding its weaknesses - it could not handle a larger scheme due to its forward-looking mechanism. Since it starts with a random policy, state evaluation is very poor at the beginning (compared to their optimal value), and the improvement becomes slow and inefficient throughout iterations. Magnification of this problem is taking place since unlike in infinite horizon formulations, different policies are used for different time-steps.
\subsection{Algorithm 2}
Algorithm 2 was tested with two different lookahead horizons: $H=1$ and $H=3$. The extremely low run-time for $H=1$ make it the most preferable algorithm for this problem, in spite of the small increase in objective cost.\\
The large difference in run times for the two cases is due to the exponential complexity in $H$. \\
The improved version of algorithm 2, which includes sub-sampling of actions, enables a  reduction in run-time, while compromising negligible value in the overall cost. %TODO: is that sentence grammarly correct?

\subsection{Algorithm 3}
The terminal state of algorithm 2's solution is fed as an initial state to algorithm 3, Sampling count used was $N_s=50$.
\subsection{Result Comparison}
\begin{table}[!ht]
%% increase table row spacing, adjust to taste
\renewcommand{\arraystretch}{1.2}
%% if using array.sty, it might be a good idea to tweak the value of
%% \extrarowheight as needed to properly center the text within the cells
%
\caption{Experiment results of the different algorithms}
\label{tab:comparison}
\noindent
\centering
    \begin{minipage}{\linewidth} %Use the minipage environment to footnote tables
    \renewcommand\footnoterule{\vspace*{-5pt}} %to remove the horizontal rule above the table footnote
    \begin{center}
        \begin{tabular}{ |l | c | c | }
    \hline 
    Algorithm & Objective cost [\$] & Run-time [min] \\ \hline \hline
    SA in \cite{SA} & 702,379 & N/A \\ \hline
    Adaptive SA in \cite{SA} & 644,951 & 145 \\ \hline \Xhline{2\arrayrulewidth}
    \textbf{Tree Search, H=1} & \textbf{512,850} & \textbf{2.5} \\ \hline
    Tree Search, H=3 & 512,217 & 240 \\ \hline
    Sub-sampled Tree Search, H=3 & 512,850 & 85 \\ \hline
    Back Sweep & 511,500 & 60 \\ \hline

  \end{tabular}
        \end{center}
    \end{minipage}
\end{table}
Table \ref{tab:comparison} summarizes the experiment's results. The solutions we obtain are very similar to each other, all around \$512,000 for operation cost. \\
Algorithm 2 produces a 27\% improvement in objective value compared to the state-of-the-art algorithm presented in \cite{SA}, which achieved a minimal objective of \$644,951, with only 2.5 minutes of running time, compared to 2.5 hours in \cite{SA}. 

%\begin{figure}[!ht]
%\centering
%\includegraphics[scale=0.5]{comparison}
%% where an .eps filename suffix will be assumed under latex,
%% and a .pdf suffix will be assumed for pdflatex; or what has been declared
%% via \DeclareGraphicsExtensions.
%\caption{Objective value of the different algorithms}
%\label{fig:comparison}
%\end{figure}

\section{Summary}
In this paper we introduced three algorithms from the field of reinforcement learning, one of them novel. We formulated the unit commitment problem as a Markov Decision Process and solved it using the three algorithms (successfully with two).\\
The superior results in the experiments section lead us to believe that modelling the UC problem as an MDP is highly advantageous over other existing methods, which were mentioned in the introduction section.\\
An additional significant improvement is the option of an immediate extension for a stochastic environment, which include consideration of uncertainties. Demand, generation capacity, and generation costs can be easily modelled as stochastic by setting the appropriate probabilistic transition kernel and reward function in our existing MDP model. The algorithms presented in the paper need not change for obtaining a solution for such a probabilistic version. This transition to an uncertain formulation might be very challenging \cite{RUC1,RUC2}, or even impossible when using other optimization methods. \\
% Current and future research in the field will include consideration of uncertainty in generation capacity and loads, which we believe can be easily handled using the proposed algorithms.\\
We intend to test our algorithms under such uncertainty conditions, and possibly to change the formulation in order to obtain a risk-averse strategy for the unit commitment. This can be done by including a risk criterion in the objective, that will take into account contingencies, shut-downs, and load shedding costs while considering the probabilities of those events to happen.

\end{document}